# Biography-based Robot Games for Older Adults

Benedetta Catricalà, Miriam Ledda, Marco Manca, Fabio Paternò, Carmen Santoro, Eleonora Zedda

CNR-ISTI, HIIS Laboratory, Pisa, Italy – {benedetta.catricala, miriam.ledda, marco.manca, fabio.paterno, carmen.santoro}@isti.cnr.it

One issue in aging is how to stimulate the cognitive skills of older adults. One way to address it is the use of serious games delivered through humanoid robots, to provide engaging ways to perform exercises to train memory, attention, processing, and planning activities. We present an approach in which a humanoid robot, by using various modalities, propose the games in a way personalised to specific individuals' experiences using their personal memories associated with facts and events that occurred in older adults' life. This personalization can increase their interest and engagement, and thus potentially reduce the cognitive training drop-out.

CCS CONCEPTS • Insert your first CCS term here • Insert your second CCS term here • Insert your third CCS term here

## 1 INTRODUCTION

The increasing number of older adults implies an increasing need for their physical, social, and cognitive assistance. Indeed, aging has a considerable impact on the health of older adults in terms of cognitive and physical impairments, which influence the abilities to complete and perform basic activities of daily living, such as cooking, shopping, managing the home, bathing, and dressing. Nowadays, a large proportion of cognitive assistance is provided by informal caregivers, usually family members. These caregivers often experience a negative impact on their psychological, emotional, and physical well-being due to the high workload [2]. Given the high health care expenditure at older ages, and their effects on family caregivers, new technologies to assist older adults with cognitive impairments are urgently needed. Non-pharmacological interventions, such as physical training, cognitive training, and social stimulation activities have been used to mitigate the cognitive decline by maintaining or improving cognitive abilities, social well-being, and quality of life of older adults [2, 3]. However, traditional interventions require experienced instructors who may be unavailable. Assistive technologies can provide useful support to address this problem. They are technologies that aim to assist different types of users during their rehabilitation. They can help older adults maintain their independence during daily routines and can also be an important instrument during their rehabilitation [11].

In recent years, humanoid robots have increased their similarity to human behaviour starting from gestures and facial expressions to the ability to understand questions and provide answers. Thanks to such humanlike characteristics, the interaction between people and robots is becoming more natural. The behaviour of such robots can also be personalised through end-user development approaches, such as trigger-action rules and associated support [6]. A recent literature review [10] indicates that the humanoid robot is an interactive technology still not sufficiently investigated for supporting the cognitive stimulation of older adults. In this paper, we present a novel approach based on a Pepper humanoid robot, which exploits serious games for cognitive stimulation of older adults. A humanoid robot is a system that can employ different interaction strategies, such as verbal and non-verbal communication, facial expressions, communicative gestures, and can detect the surrounding context by using various sensors (tactile sensors, camera, microphones). These capabilities are essential to creating social and emotional interaction with users to increase their acceptability and user's engagement, which may increase the possibility of reaching the goal of assistance in less time and with better results [2].

Using robots to support and assist patients can be a valuable tool to help them during their cognitive training. In such context, digital cognitive training through serious games may potentially benefit those with cognitive impairments more than traditional training due to enhanced motivation and engagement. In literature, different studies show how digital games can obtain positive results in helping seniors improve their cognitive abilities compared to traditional training [8]. Since older adults are varied in terms of preferences, interests, and abilities, it is important to propose serious games for cognitive training that are able to personalise, and thus be more relevant for them. Combining a humanoid robot and a set of personalised serious games can be a solution to obtain measurable progress in cognitive functions and stimulate the user to continue the training [13]. Personalised serious games for cognitive intervention have been explored with mobile apps [15] but have not been investigated with humanoid robots. We aim to offer novel digital training through serious games designed using personally relevant material from older adults' life. They will be based on elements associated with their biography, thus making interactions personalised, relevant, and more engaging.

## 2 THE SERENI APPROACH

The psychological well-being of older adults may be affected by some age-related conditions, such as approaching death, loss of family members, and reduced autonomy. A meta-analysis [2] indicates that the practice of life review (discussing what a memory means), even more than reminiscence (describing a memory itself), is a good instrument for improving the psychological well-being of older adults and that its effect sizes are comparable to those of cognitive-behavioural therapy. Serrano et al. [11] found that the practice of autobiographical memory improved the mood of the elderly by improving their life satisfaction. Furthermore, Damianakis et al. [4] report that interventions that contextualise history, personality, and life experiences can contribute to improving both communication and social interactions between family members and between family members and formal caregivers. Based on previous experiences [7], we have started the development of a new prototype in which the serious games installed on the humanoid robot will motivate older adults by engaging them in playful situations that draw on their personal memories, with which they can interact. Indeed, such serious games are designed to use personally relevant material and events from older adults' life. Specifically, the games are based on elements associated with the biography of the users (mainly taken from from their youth), thus making interactions more relevant and more likely to keep them engaged while enhancing their well-being.

According to such motivations, we have designed the SERENI platform to deliver serious games using personally relevant material from older adults' life through a humanoid robot. It aims to stimulate cognitive functions through play sessions, which should last 15-20 minutes. The exercises should be useful for making the participants think and reason before providing the correct answer. The platform can be a solution for day-care centres where older adults with mild cognitive impairments can go to perform relevant exercises. On the one hand the older adults, by interacting with the biographical app, provide relevant biographical data that are mainly used to customise the games, which thereby will be highly personalised for them. On the other hand, seniors will also interact with the games to stimulate their cognitive abilities. The data produced during the interactive sessions will be exploited to improve the adaptation of the game itself (according to the data gathered in previous game sessions) and also to feed the associated analytics services.

The SERENI platform is based on a modular architecture allowing the deployment of multimodal serious cognitive games on a humanoid robot. Thanks to its human-like appearance and behaviour, it can stimulate interest and engagement from seniors that would be more difficult with other types of smaller and more limited robots can stimulate interest and engagement from seniors that would be more difficult with other types of smaller and more limited robots, thanks to its human-like appearance and behaviour. The platform is based on various components. The first one is the Remind App, a responsive multimodal Web application to collect memories from older adults and their relatives. The memories can be entered both through graphical and vocal interaction. Biographical information is exploited in a group of games that aim to stimulate and train various cognitive resources in older adults (memory, attention, planning). The platform is also able to store data regarding user performance (i.e. when and for how long the user played with a given game, number of errors in a session, type of games played). In the resulting environment, the humanoid robots will serve as personal trainers, proposing exercises and communicating through various modalities, and challenging users in cognitive games relevant to their daily life (e.g. by remembering past events or names of family members and friends). The solution aims to allow caregivers to configure the exercises and choose the most suitable games to stimulate the cognitive skills of users and enhance their experience. Caregivers can also interact with an Analytics tool, to have both overview and detailed information regarding user performance and state. For this goal, the games include a custom tracking system, which tracks the data about user performance and other game analytics data (such as time, number of errors, pass/fail, score, completion level, etc.).

To facilitate entering the memories through the responsive Web application (Remind) developed to collect older adults' memories, we thought it was useful to categorise the biographical aspect, also because different types of memories need different types of questions for being entered. Based on our previous experiences in projects in the Ambient Assisted Living area and informal discussions with relevant stakeholders, in the first version of the app we identified a first set of memory categories: Music, Events, Games, Places, Food, and Hobbies. We then decided to carry out an empirical validation of such classification with the target audience, by proposing to people aged 65+ a questionnaire (in Italian), composed of three parts. The initial part was dedicated to demographic information, in the second part they were asked to freely indicate at least four categories that they deemed particularly relevant to classify their personal memories, and to select the categories they find relevant within Food, Events, Family, Travels, Music, Hobby, Work, Love/Friendship, Study, Health. Then, they had to rate on a scale from 1 to 5 the relevance of the categories used in the initial version of the games (Locations, Games, Hobby, Food, Music, and Personal Events), with the possibility of indicating the category they would add or remove.

The questionnaire was completed in a paper form by 50 people (23 males and 27 females) aged between 65 and 84 years (Mean: 72, SD: 5,09). 40% have a higher education, 38% have a degree. 86% very familiar with electronic devices such as smartphones, tablets and PCs; the remaining 14% only use smartphones mainly out of necessity. 80% indicated "Family" as the most representative category for their memories; the examples proposed concern the birth of children and grandchildren, the memory of parents and grandparents and the childhood home. 40% of them indicated "Work", in particular the first experiences and satisfactions during their career. 50% cited "Affections", and provided examples such as meeting their first love, childhood friendships, and events such as engagement and marriage.

Participants were also asked to indicate, among the initially proposed categories, those most relevant to them. Participants rated each category on a scale of 1 (Not Relevant) to 5 (Relevant). In particular, 54% rated "Hobbies" as Not Relevant (scores < 3); 40% rated "Food" as Not Relevant (scores < 3); 74% rated "Music" as Relevant (scores > 3); 86% rated "Places" as Relevant (scores > 3); 98% rated "Events" as Relevant (scores > 3); 68% rated "Games" as Relevant (scores > 3). The most relevant category among those proposed by us was "Events" (avg score = 4,7): it was considered very versatile by users, as it allows the inclusion of different types of memories. The least relevant categories were Hobbies and Food. Hobbies received an average score of about 2,6: the main criticism concerned the category's name which was judged not adequate compared to other proposals. As possible replacements, terms such as "leisure" or "entertainment" were suggested. Participants also showed very low interest in the Food category, as most of them said that it did not significantly impact on their life experiences. In conclusion, the most significant memories concern the dearest affections and memorable events in life (i.e. graduation, marriage, children birth). Of the six categories proposed, Music and Personal Events aroused the greatest interest. Thus, in the new version of the Remind app we introduced the Affections category and removed the Food one. In the end, the categories selected were: Affections, Events, Games, Hobbies, Places, Music.

At the beginning of the interaction with the Web Remind application, users are asked whether they want to enter a new memory or review those previously entered. After selecting a memory category, the user can provide the associated information associated with the specific memory. For example, for entering a memory related to a particular event in life the user indicates a name for the event and provide a description, which can be entered either vocally or by keyboard. The users can also indicate their age when such event occurred, and optionally provide an image associated with it. In the case of a memory in the Hobby category, the user can also provide a list of activities required by the hobby. All such information can then be used by the games provided by the Pepper robot for specific exercises. In general it is not necessary that the older adults directly enter the memories, to facilitate the process they can tell them to some formal or informal caregiver, who can also help them in specifying relevant memories. The Pepper application presents various exercises useful for making the participants think and reason to provide the correct answer. An initial set of five games have been identified:

- **Memory completion**. Pepper presents a memory with a missing detail, which the user should select from some elements (if the answer is correct, the memory is re-read to the senior). For example: "when I was 12 years old I used to spend summer time in…" and the robot shows three possible options: Marina di Pisa, Tirrenia, San Vincenzo or Castiglioncello;) or "I used to listen to that singer when I travelled by car with my father" with possible answers: Modugno, Morandi, Celentano, Guccini;
- **Activities ordering**. It is only applied to Hobby: a set of activities presented in an unordered list should be put in the right order by the user (this can stimulate executive functions and procedural memory);
- **Memory association**. In this game, 3-4 memories are briefly listed as well as some details: users have to connect each memory with the corresponding detail, for example associating song titles to the corresponding singers (to stimulate attention and memory);
- **Memory-related event** question. The user has to guess an event that happened in the same year of the memory: the robot asks the user to select that event from a list of possible events. For example: what happened in the same year you got married (1945)? Possible answers: "the end of second world war", "the first man on the moon", "women gain the right to vote in Italy"? (useful to stimulate long-term memory).
- **Music game**, the robot plays the initial part of a song popular at the time of the memories and the user has to guess its singer or title. In general, music has a positive effect on the users engagement, and in this case music related to their memories is proposed.

In a session at the beginning the robot asks the name of the user, then through such information it retrieves the memories that the user entered, which are available from the biography application backend through a restful service and transmitted in JSON format. The memories arrive in the robot with the indication of the corresponding category, which is useful to determine how to exploit them in the various exercises. In the case of a missing detail in the Memory completion exercise, the robot proposes a memory and a list of possible missing details derived from that user's memories. For the memory-related event exercises, the list of options in terms of real events are taken by external services. The activities ordering exercise refers only to the Hobby category because only in that case users are asked to enter the steps required to perform the hobby. Thus, users can first select the type of game they want to play, and then they have the opportunity to perform the associated exercises, with personalised content.

## 3 CONCLUSIONS AND FUTURE WORK

In this paper, we introduce a novel approach to personalising serious games for cognitive stimulation of older adults delivered through a humanoid Pepper robot. It is based on a multimodal Web app to collect memories of older adults, and then such content is exploited in a set of games aiming to stimulate several cognitive resources of seniors. We have collected memories from 16 older adults (65+) with MCI and in next weeks we will carry out a trial in which they will be asked to interact with both the version of the games exploiting personal memories and another version with standard content in a within-subjects study so that we can assess the impact of the biography-based personalization.

## ACKNOWLEDGMENTS

This work is partly supported by the CNR project SERENI https://hiis.isti.cnr.it/sereni/index.html